# Healthcare NER Models Using Language Model Pretraining

Empirical Evaluation of Healthcare NER Model Performance with Limited Training Data

Amogh Kamat Tarcar [*]
Persistent Systems Limited, Goa, India
amogh_tarcar@persistent.com

Aashis Tiwari
Persistent Systems Limited, Pune, India
aashis_tiwari@persistent.com

Dattaraj Rao
Persistent Systems Limited, Goa, India
dattaraj_rao@persistent.com

Vineet Naique Dhaimodker
National Institute of Technology, Goa, India
abcdvineet27@gmail.com

Penjo Rebelo
National Institute of Technology, Goa, India
rpenjo0007@gmail.com

Rahul Desai
National Institute of Technology, Goa, India
rahulgdesai1998@gmail.com

## ABSTRACT

In this paper, we present our approach to extracting structured information from unstructured Electronic Health Records (EHR) [2] which can be used to, for example, study adverse drug reactions in patients due to chemicals in their products. Our solution uses a combination of Natural Language Processing (NLP) techniques and a web-based annotation tool to optimize the performance of a custom Named Entity Recognition (NER) [1] model trained on a limited amount of EHR training data. This work was presented at the first Health Search and Data Mining Workshop (HSDM 2020) [26].

We showcase a combination of tools and techniques leveraging the recent advancements in NLP aimed at targeting domain shifts by applying transfer learning and language model pre-training techniques [3]. We present a comparison of our technique to the current popular approaches and show the effective increase in performance of the NER model and the reduction in time to annotate data. A key observation of the results presented is that the F1 score of model (0.734) trained with our approach with just 50% of available training data outperforms the F1 score of the blank spaCy model without language model component (0.704) trained with 100% of the available training data.

We also demonstrate an annotation tool to minimize domain expert time and the manual effort required to generate such a training dataset. Further, we plan to release the annotated dataset as well as the pre-trained model to the community to further research in medical health records.

## 1. INTRODUCTION

Extracting structured information from unstructured text such as EHRs and medical literature has always been a challenging task. Recent advancements in machine learning take advantage of the large text corpora available in scientific literature as well as medical and pharmaceutical web sites and train systems which can be leveraged for several NLP tasks ranging from text mining to question answering. Along with progress in the research space, there has been significant progress in the libraries and tools available for industry use.

The specific problem we focused on was extracting adverse drug reactions from EHRs using NER. The solution required us to extract key entities such as prescribed drugs with dosage and the symptoms and diseases mentioned in the EHRs. The extracted entities would be processed further downstream to link the entities and leverage dictionary-based techniques for flagging any symptoms which could potentially be adverse drug reactions of the prescribed medicines.

A crucial component in our devised solution employed a custom NER model for extracting key entities from EHRs. The state-of-the-art Named Entity Recognition models built using deep learning techniques [13] extract entities from text sentences by not only identifying the keywords or linguistic shape of entities, but also by leveraging the context of the entity in the sentence. Furthermore, with language model pre-trained embeddings, the NER models leverage the proximity of other words which appear along with the entity in domain specific literature.

One of the key challenges in training NLP based models is the availability of reasonable-sized, high-quality annotated datasets. Further, in a typical industrial setting, the relative difficulty in garnering significant domain expert time, and the lack of tools and techniques for effective annotation along with the ability to review such annotations to minimize human errors , affects research and benchmarking new learning techniques and algorithms.

Additionally, models like NER often need significant amount of data to generalize well to a vocabulary and language domain. Such vast amounts of training data are often unavailable or difficult to manufacture or synthesize.To bridge the gap between academic developments and industrial requirements, we designed a series of experiments employing transfer learning from pre-trained models while working with a comparatively smaller dataset.

Transfer learning techniques [3] are largely successful in the image domain and are advancing steadily in natural language domain with the availability of pre-trained language embeddings and pre-trained models.

In this paper we present findings of our experiments to solve the industrial problem of training NER models with limited data using spaCy [7] , a state-of-the-art industrial strength natural language processing package, along with the latest techniques in transfer learning.

The paper is organized as follows. Section 2 describes the motivation for our experiments followed by Section 3 discusses the problem and our solution, both in algorithmic and implementation terms, and evaluates the results produced by our solution. Section 4 discusses the results and Section 5 concludes and suggests directions for future work.

## 2. MOTIVATION

Recent advancements in NLP also known as the ImageNet moment in NLP [3], have shown significant improvements in many NLP tasks using transfer learning. Language models like ELMo [4] and BERT [5] have shown the effect of language model pre-training on downstream NLP tasks. Language models are capable of adjusting to changes in the textual domain with a process of fine-tuning. Also, in this self-supervised learning scenario, there is an implicit annotation in sentences, i.e.  predicts the next token (word) given a sequence of tokens appearing earlier in the sequence. Given all this, we can adjust to a new domain-specific vocabulary with very little training time and almost no supervision.

NER aimed at detecting and identifying entity classes in text can help in extracting structured information and assisting upstream user experiences. The applicability of NER models are widespread, ranging from identifying dates and cities in chatbots to open domain question answering.

Using task-specific annotation tools can minimize the time to generate high-quality annotated datasets for training models. The traditional process of annotating data is slow, but fundamental to most NLP models. It often acts as a hindrance in evaluating and benchmarking multiple models, as well as in parameter tuning of models. Many tools such as Doccano [6] exist in the open-source community that help in solving this problem. We developed an in-house tool which we could customize for speeding up the annotation process.

## 3. USE CASE DETAILS

Studying adverse reactions due to chemicals in a drug on the patient is central to drug development in healthcare. Pharmacovigilance (PV) [21] as described by WHO, is defined as "the science and activities relating to the detection, assessment, understanding and prevention of adverse effects or any other drug-related problem." Pharmaceutical companies often want to understand the conditions and pre-conditions under which a drug might have an adverse reaction on a patient. This would help in research and studies of the drugs and also reduce or prevent risks of any harm to the patient.

Co-occurrence of disease and chemicals in an EHR of a patient is useful in studies and research for most pharmaceutical companies. However, EHRs are unstructured data and additional processing is required to extract structured information such as named entities of interest. Such extraction can lead to significant savings of manual labor and minimizing the time taken to get a new drug to market.

We developed custom healthcare NER models to extract phrases related to (pharmaceutical) chemicals with dosage, diseases and symptoms from EHRs. As the entities were specific to the domain text, an in-house annotated dataset was created using our custom-built annotation tool. A number of experiments were designed and executed for training custom NER models on annotated data from base models (spaCy[7] and scispaCy[8]) using transfer learning. Section 3.1 describes the dataset preparation followed by Section 3.2 which presents an architecture overview. Section 3.3 presents experiment details and Section 3.4 describes the results obtained.

### 3.1. DATASET PREPARATION
We created a domain-specific corpus by collating publicly available sample medical notes and drug public assessment reports from European Medical Agency (EMA )[9] and Sample Medical Transcripts [10].

- A custom annotated dataset was created in-house specifically for the four entities: Chemical, Disease, Symptom and Dosage.
- A text corpus containing domain specific vocabulary was created by utilising text from 2300 sample notes from the Medical Transcripts Samples site and 100 FAQ sections from the EMA site.

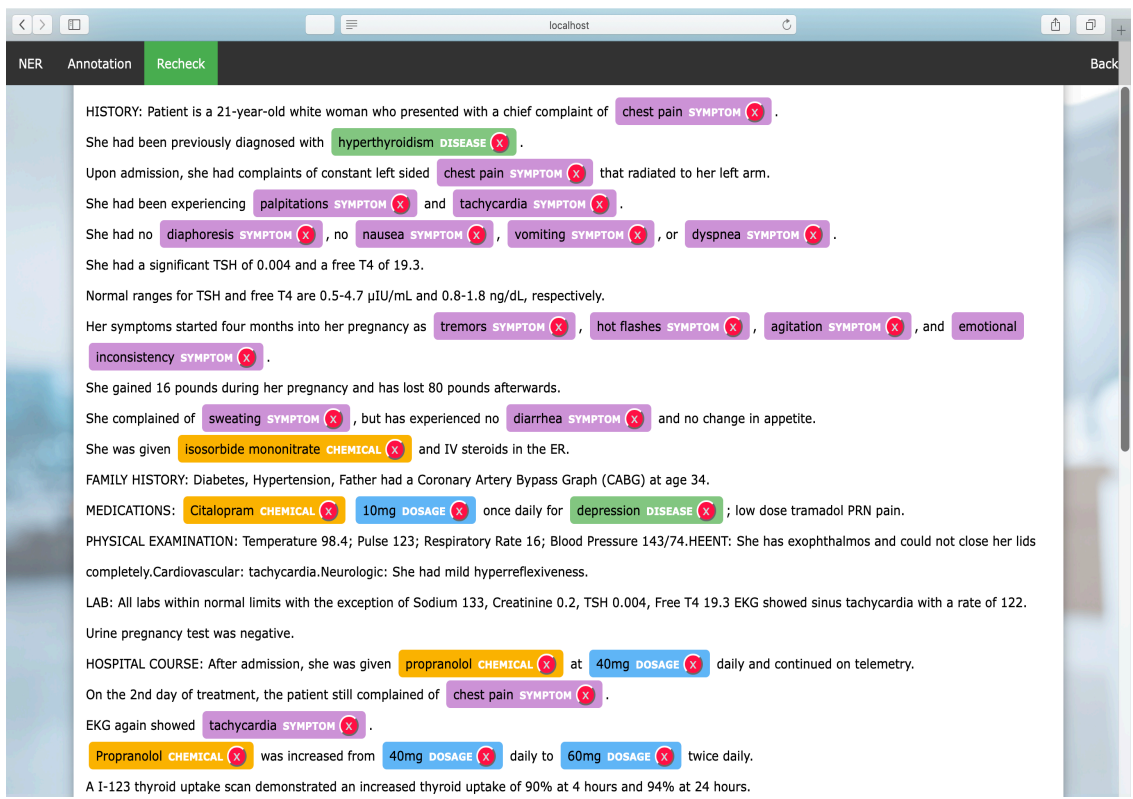

**Figure 1: In-house Annotation Tool**

For annotating data, a custom-built web browser-based tool was used. Figure 1 displays a screen shot of the annotation tool. As seen in the figure, the tool works with text files and the user provides annotations using mouse and keyboard inputs. After marking the required span of text using the mouse, the user can use keyboard keys to annotate the selected span. For example, the 'S' key on the keyboard represents the Symptom entity. On providing inputs, the tool highlights the span with a specific color for each entity, and also adds an entity name on the screen with a cross mark to make corrections. The tool also has a recheck functionality to enable the reviewer to reexamine annotations.

After initial annotations (Around 100 occurrences of each entity), we utilized the annotated data to train the spaCy [7] NER model and leveraged it to identify named entities in new text files to accelerate the annotation process.

The annotated dataset was randomized and split into 80% for training and 20% for testing. As the training data for spacy follows a pattern of sentence and entity tuples there is no overlap between sentences split into training and test dataset. The training data was further split into smaller sets ranging from 50% to 100% of the data, in 10% increments. Table 1 presents the statistics of the annotated data. It tabulates the counts of annotated sentences as well entity wise counts.

**Table 1. Dataset Description**

| Dataset | Total Sentences | Entity Counts | | | |
|---|---|---|---|---|---|
| | | Chemical | Disease | Symptom | Dosage |
| All Annotated Data | 4212 | 1194 | 929 | 1922 | 290 |
| Train Data | 2948 | 908 | 638 | 1351 | 207 |
| Test Data | 1264 | 286 | 291 | 571 | 83 |

### 3.2. ARCHITECTURE OVERVIEW

Figure 2 presents our solution architecture which includes four key components. The first component comprises of Python scripts to fetch and collate medical notes text from the Sample Medical Transcripts [10] site. The second component is the in-house Annotation tool which is used by domain experts to annotate notes. The annotation tool is a web application with a Python Flask- [14] based REST API as the backend. The annotation tool processes document annotations and outputs annotated data in the format required to train spaCy models. The third component consists of a Python module which utilizes the spaCy pre-train feature for language model pre-training. The fourth component is a Python module built using spaCy which consumes the annotated data, the spaCy models and the pre-trained vector to performs model training, to produce custom NER models.

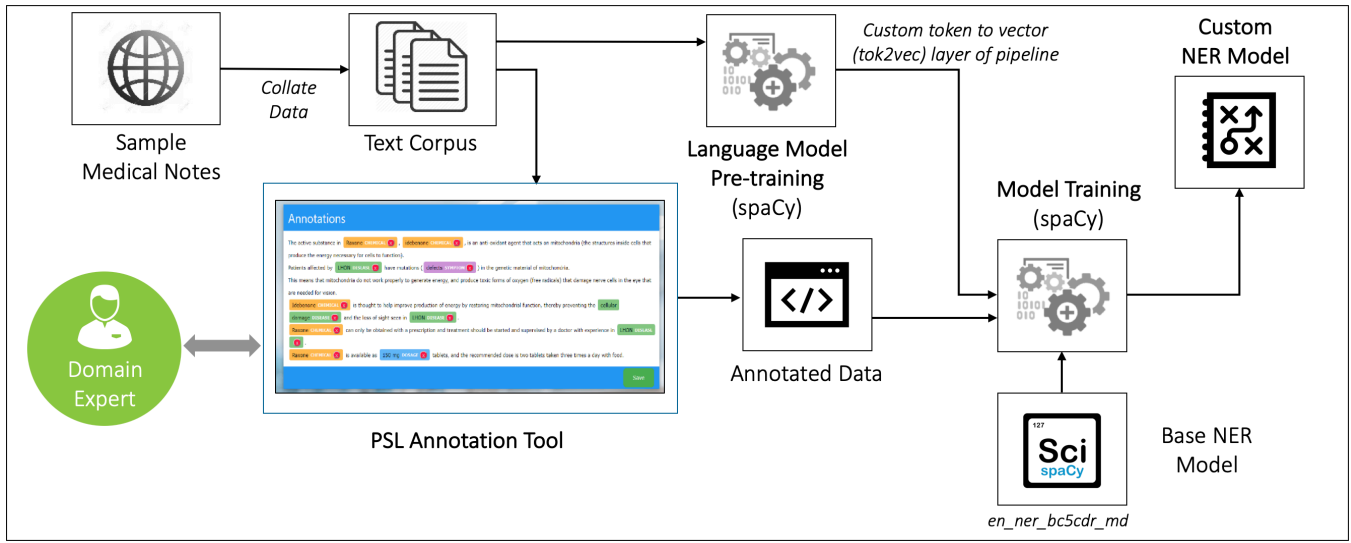

**Figure 2: Architectural Overview of Experiment Setup**

### 3.3. EXPERIMENTS

The spaCy library provides a variety of tools for fast text processing and is developed as a modular pipeline. The library parses text to create a custom spaCy data structure which is then passed through orchestrated components of the pipeline for further processing. The components of the pipeline are highly customizable for efficient execution of NLP tasks such as text categorization, POS tagging , named entity recognition etc. Furthermore, these components can be individually updated for adapting to specific implementations.

For implementation of our experiments we focused onto two critical components of the spaCy pipeline namely the components which are responsible for converting string tokens to vectors and the named entity recognition component.

The NER component in spacy pipeline is a deep learning model utilizing Convolutional Neural Network (CNN) and Long Short Term Memory (LSTM) architectures. The implementation is based on the transition-based framework described by Lample et al [21].

SpaCy provides wrapper APIs to interact with this NER model as well as to improve it and add custom entities by leveraging annotated data. The spaCy training API obtains error gradients and updates the model weights using back-propagation. Using spaCy train API [15], we can train the deep learning model for NER component starting from a blank English language class model which has no learned entities or leverage a model which already contains a few trained entities. Leveraging the model which has been trained for recognizing a few overlapping entities is often beneficial in case of limited amount of training data as compared to the blank model.

The second component which we experimented with was the component in the pipeline which provides vectors for string tokens. The token to vector layer of the pipeline (tok2vec) can be customized to provide custom vectors. There are two prominent techniques for obtaining word embeddings: the classic techniques such as Word2vec[22] and Glove [23] which provides static embeddings for each word, and the dynamic word embeddings which are based on the context of the word in the sentence. Dynamic word embeddings can be obtained from language models such as ELMo [4] and BERT [5].

In order to obtain dynamic embeddings specific to the context of the text which we need the NER model to run on we need to train a separate machine learning model. The learning objective of training task is to work with the non-annotated text corpus by internally converting it into a supervised learning task by masking words from sentences and then predicting this word. This training task is implemented on raw text corpus containing a large number of domain specific words. The embeddings are then obtained from this trained model which can be leveraged as token vectors.

spaCy has implemented a deep learning implementation for obtaining dynamic word embeddings using an approximate language-modelling objective. The pretrain wrapper API [12] internally executes the training of this deep learning model given a large corpus of domain specific text data. The output of pretraining API is a domain specific dynamic embedding model.

We designed three experiments using these two key components of the spaCy NLP pipeline and trained multiple NER models using the annotated training data to obtain optimal performance on test data using the spaCy training module [15]. Our experimental set up included working with spaCy version 2.1.4 [18] on an Anaconda Distribution [16], Python 3.6.8 [17] environment running on a machine with x86_64 GNU/Linux, Intel Core Processor (Broadwell) with 16 GB RAM. The experiments can be split into three main methods.

### 3.3.1. Method 1: Blank spaCy model

We trained a blank spaCy English language model (this model has no trained entities) using annotated training data to recognize four custom entities. We did not provide any custom token to vector layer and set the API to use default execution of the spaCy NLP pipeline.

We started with utilizing only 50% of the available training data and trained 5 models (for 100 iterations with dropout rate=0.2) while increasing the training data in increments of 10%. The performance of the trained models was evaluated on the test data.

### 3.3.2. Method 2: scispaCy + Transfer Learning + Retraining

We observed that the pre-trained scispaCy model (*en_ner_bc5cdr_md*) [8] was trained on BC5CDR [11]corpus for recognizing two entities (Disease and Chemical) that overlap with our custom four entities. The BC5CDR corpus consists of 1500 PubMed articles with 4409 annotated chemicals, 5818 diseases and 3116 chemical-disease interactions [20].

As the model was already trained on medical data, we used it as a base model and applied transfer learning and retrained it using our in-house annotated data. Similar to the blank models, we trained 5 models (for 100 iterations with dropout rate=0.2) while increasing the training data from 50% to 100% of the available training data. The performance of the trained models was evaluated on the standard test data.

### 3.3.3. Method 3: scispaCy + Transfer Learning + Pre-training

In order to improve the performance of transfer learning models further, we employed a newly released spaCy package feature, that of *pre-training*. Pre-training allows us to initialize the neural network layers of spaCy's CNN layers with a custom vector layer. This custom vector can be trained by utilizing a domain specific text corpus using the spaCy library pre-training command [12].

The pre-training API spaCy has implemented a deep learning implementation for obtaining dynamic word embeddings using a Language Modelling with Approximate Outputs (LMAO) described in spaCy Language model pretraining [25].

We leveraged spaCy pre-training API and trained our custom dynamic embedding model over our domain specific text corpus. We collated our domain specific text corpus (which was created by utilizing text sentences from 2300 sample notes from Sample Medical Transcripts [10] site and 100 FAQ section texts from EMA [9] site).We provided scispaCy model *(en_ner_bc5cdr_md)* [8] embedding vectors as a seed while training.

We monitored the loss over epoch while training and it was observed that the loss gradually reduced to minimum around 95 epoch mark after which it plateaued. With our experimental set up (CPU machine with x86_64 GNU/Linux, Intel Core Processor (Broadwell), 16 GB RAM), 95 epochs of fine tuning were completed in 8 hours.

Then we used this domain specific word embedding model for the vectorization of tokens while performing transfer learning from scispaCy model (*en_ner_bc5cdr_md*) [8] using our annotated data.

We trained five models (for 100 iterations with dropout rate=0.2) similar to the models developed in earlier methods. With our experimental set up (CPU machine with x86_64 GNU/Linux, Intel Core Processor (Broadwell), 16 GB RAM), 100 iterations of training required for each model were completed in 48 mins.

### 3.4. RESULTS

Table 2 captures the observed overall NER model performance on test data for the conducted experiments.

**Table 2. Performance of Trained Models on Test Data**

| Percentage of Training Data Used | Model Name | Performance on Test Data | | |
|---|---|---|---|---|
| | | Precision | Recall | F1-Score |
| 50 % | Blank | 0.607 | 0.539 | 0.571 |
| | Retrained scispaCy | 0.682 | 0.719 | 0.700 |
| | Retrained scispaCy with pre-training | 0.711 | 0.759 | 0.734 |
| 60 % | Blank | 0.647 | 0.569 | 0.605 |
| | Retrained scispaCy | 0.714 | 0.728 | 0.721 |
| | Retrained scispaCy with pre-training | 0.740 | 0.758 | 0.749 |
| 70% | Blank | 0.688 | 0.611 | 0.647 |
| | Retrained scispaCy | 0.744 | 0.752 | 0.748 |
| | Retrained scispaCy with pre-training | 0.753 | 0.747 | 0.750 |
| 80% | Blank | 0.689 | 0.646 | 0.667 |
| | Retrained scispaCy | 0.755 | 0.741 | 0.748 |
| | Retrained scispaCy with pre-training | 0.757 | 0.778 | 0.767 |
| 90% | Blank | 0.696 | 0.662 | 0.679 |
| | Retrained scispaCy | 0.747 | 0.743 | 0.745 |
| | Retrained scispaCy with pre-training | 0.754 | 0.761 | 0.757 |
| 100% | Blank | 0.724 | 0.685 | 0.704 |
| | Retrained scispaCy | 0.755 | 0.743 | 0.749 |
| | Retrained scispaCy with pre-training | 0.776 | 0.794 | 0.785 |

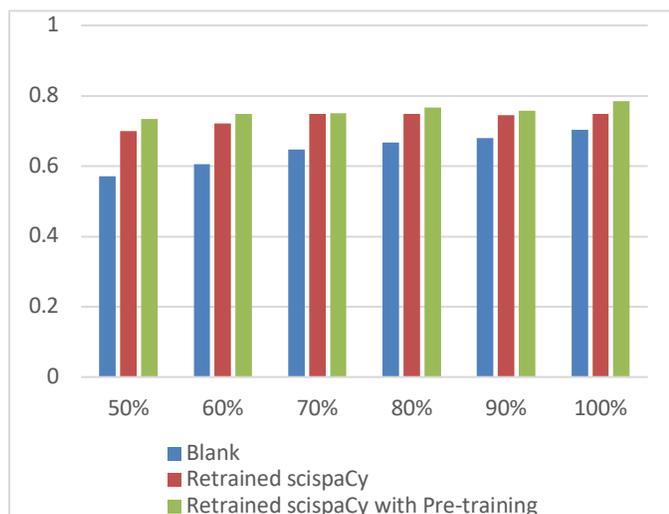

**Figure 3: F1 Scores on Test Data While Increasing Training Data**

For each trained model overall NER evaluation metrics were recorded which includes Precision, Recall and F1 Score [19]. Figure 3 presents bar chart representation of the observed overall F1 scores as mentioned in Table 2, across the three methods named as Blank, Retrained scispaCy and Retrained scispaCy with pre-training while progressively increasing training data.

As observed in Figure 3, there is generally a steady increase in F1 score with an increase in available training data. The gain between blank model and scispaCy derived models is prominent along with a steady gain visible between the two scispaCy derived models.

Table 3 presents the entity-wise F1 scores [19] of the models trained using 100% training data using the three methods. As observed in Table 3, the F1 scores of the model derived from scispaCy with pre-training are consistently higher than the other models across the entities.

**Table 3. Entity-wise F1 Scores of Trained Models on Test Data with 100% Training Data**

| Label | Blank | Retrained scispaCy | Retrained scispaCy with Pre-training |
|---|---|---|---|
| CHEMICAL | 0.790 | 0.842 | 0.860 |
| DISEASE | 0.690 | 0.785 | 0.809 |
| SYMPTOM | 0.637 | 0.719 | 0.723 |
| DOSAGE | 0.815 | 0.838 | 0.878 |

## 4. DISCUSSION

As observed in the results, with progressive increase in availability of training data, the performance of the models on test data steadily increases. A clear gain is observed between the blank model and the model based on scispaCy pre-trained model. This gain can be attributed to the overlap of entities between the custom model and the scispaCy model. Furthermore, performance gains are observed when using a pre-training vector customized to the domain vocabulary used in the medical reports.

A key observation of the results presented is that the F1 score of the scispaCy + pre-trained model trained with just 50% of available training data (0.734, as observed in Table 2 in Section 3.4) outperforms the F1 score of the blank spaCy model trained with 100% of the available training data (0.704, as observed in Table 2 in Section3.4)

The final performance of custom NER model was evaluated on the test data set. The overall F1 score of our recommended NER model which was derived from scispaCy (*en_ner_bc5cdr_md*) [8] using Method 3 with custom pre-trained vector was 0.785 as observed in Table 2 in Section 3.4.

## 5. CONCLUSION

Our experiments present empirical results which corroborate the hypothesis that transfer learning delivers clear benefits while working with even a limited amount of training data. A key observation of the results presented is that the F1 score of a model trained with our approach with just 50% of available training data (0.734) outperforms the F1 score of the blank spaCy model (0.704) trained with 100% of the available training data. Clearly, leveraging pre-trained models with partial overlap with the entities provides significant benefits.

In future work, we plan to increase the number of entities and experiment with how the number of entities affect performance of the trained models. We also plan to release our pre-trained model with pharmacology domain entities that can be used for multiple applications.

Our approach to the problem using a custom annotation tool and pre-training techniques can be utilized and extended to multiple NLP problems, such as Machine Comprehension, FAQ-based Question-Answering, Text Summarization etc. The techniques are application domain-agnostic and can be applied to any industrial vertical such as but not limited to: Banking, Insurance, Pharma, Healthcare etc., where domain expertise is required.